**Title**
AI system for fetal ultrasound in low-resource settings

**Authors**
Ryan G. Gomes[a]; Bellington Vwalika[bc]; Chace Lee[a]; Angelica Willis[a]; Marcin Sieniek[a]; Joan T. Price[cd]; Christina Chen[a]; Margaret P. Kasaro[bd]; James A. Taylor[a]; Elizabeth M. Stringer[c]; Scott Mayer McKinney[a]; Ntazana Sindano[d]; George E. Dahl[a]; William Goodnight, III[b]; Justin Gilmer[a]; Benjamin H. Chi[cd]; Charles Lau[a]; Terry Spitz[a]; T Saensuksopa[a]; Kris Liu[a]; Jonny Wong[a]; Rory Pilgrim[a]; Akib Uddin[a]; Greg Corrado[a]; Lily Peng[a]; Katherine Chou[a]; Daniel Tse[a]; Jeffrey S. A. Stringer[cd]; Shravya Shetty[a]

Ryan G. Gomes, Bellington Vwalika, Chace Lee, and Angelica Willis contributed equally to this work.
Jeffrey S. A. Stringer and Shravya Shetty jointly supervised this work.

  a. Google Health, Palo Alto, CA, USA
  b. Department of Obstetrics and Gynaecology; University of Zambia School of Medicine, Lusaka, Zambia
  c. Department of Obstetrics and Gynecology; University of North Carolina School of Medicine, Chapel Hill, NC, USA
  d. UNC Global Projects – Zambia, LLC; Lusaka, Zambia

**Abstract**

Despite considerable progress in maternal healthcare, maternal and perinatal deaths remain high in low-to-middle income countries. Fetal ultrasound is an important component of antenatal care, but shortage of adequately trained healthcare workers has limited its adoption. We developed and validated an artificial intelligence (AI) system that uses novice-acquired "blind sweep" ultrasound videos to estimate gestational age (GA) and fetal malpresentation. We further addressed obstacles that may be encountered in low-resourced settings. Using a simplified sweep protocol with real-time AI feedback on sweep quality, we have demonstrated the generalization of model performance to minimally trained novice ultrasound operators using low cost ultrasound devices with on-device AI integration. The GA model was non-inferior to standard fetal biometry estimates with as few as two sweeps, and the fetal malpresentation model had high AUC-ROCs across operators and devices. Our AI models have the potential to assist in upleveling the capabilities of lightly trained ultrasound operators in low resource settings.

**Introduction**

Despite considerable progress in maternal healthcare in recent decades, maternal and perinatal deaths remain high with 295,000 maternal deaths during and following pregnancy and 2.4 million neonatal deaths each year. The majority of these deaths occur in low-to-middle-income countries (LMICs).[1–3] The lack of antenatal care and limited access to facilities that can provide lifesaving treatment for the mother, fetus and newborn contribute to inequities in quality of care and outcomes in these regions.[4,5]

Obstetric ultrasound is an important component of quality antenatal care. The WHO recommends one routine early ultrasound scan for all pregnant women, but up to 50% of women in developing countries receive no ultrasound screening during pregnancy.[6] Fetal ultrasounds can be used to estimate gestational age (GA), which is critical in scheduling and planning for screening tests throughout pregnancy and interventions for pregnancy complications such as preeclampsia and preterm labor. Fetal ultrasounds later in pregnancy can also be used to diagnose fetal malpresentation, which affects up to 3-4% of pregnancies at term and is associated with trauma-related injury during birth, perinatal mortality, and maternal morbidity.[7–11]

Though ultrasound devices have traditionally been costly, the recent commercial availability of low-cost, battery powered handheld devices could greatly expand access.[12,13,14] However, current ultrasound training programs require months of supervised evaluation as well as indefinite continuing education visits for quality assurance.[13–18] To address these barriers, prior studies have introduced a protocol where fetal ultrasounds can be acquired by minimally trained operators via a "blind sweep" protocol, consisting of 6 predefined freehand sweeps over the abdomen.[19–23]

In this study, we used two prospectively collected fetal ultrasound datasets to estimate gestational age and fetal malpresentation while demonstrating key considerations for use by novice users in LMICs: a) validating that it is possible to build blind sweep GA and fetal malpresentation models that run in real-time on mobile devices; b) evaluating generalization of these models to minimally trained ultrasound operators and low cost ultrasound devices; c) describing a modified 2-sweep blind sweep protocol to simplify novice acquisition; d) adding feedback scores to provide real-time information on sweep quality.

**Blind sweep procedure**
Blind sweep ultrasounds consisted of a fixed number of predefined freehand ultrasound sweeps over the gravid abdomen. Certified sonographers completed up to 15 sweeps. Novice operators ("novices"), with 8 hours of blind sweep ultrasound acquisition training, completed 6 sweeps. Evaluation of both sonographers and novices was limited to a set of 6 sweeps - 3 vertical and 3 horizontal sweeps (Figure 1B).

**Fetal Age Machine Learning Initiative (FAMLI) and Novice User Study Datasets**
Data was analyzed from the Fetal Age Machine Learning Initiative cohort, which collected ultrasound data from study sites at Chapel Hill, NC (USA) and the Novice User Study collected from Lusaka, Zambia (Figure 1A).[24] The goal of this prospectively collected dataset was to empower development of technology to estimate gestational age.[25] Data collection occurred between September 2018 and June 2021. All study participants provided written informed consent, and the research was approved by the UNC institutional review board and the biomedical research ethics committee at the University of Zambia. Studies also included standard clinical assessments of GA and fetal malpresentation performed by a trained sonographer.[26] Blind sweep data were collected with standard ultrasound devices (SonoSite M-Turbo or GE Voluson) as well as a low cost portable ultrasound device (ButterflyIQ).

Evaluation was performed on the FAMLI (sonographer-acquired) and Novice User Study (novice-acquired) datasets. Test sets consisted of patients independent of those used for AI development (Figure 1A). For our GA model evaluation, the primary FAMLI test set comprised 407 women in 657 study visits in the USA. A second test set, "Novice User Study" included 114 participants in 140 study visits in Zambia. Novice blind sweep studies were exclusively performed at Zambian sites. Sweeps collected with standard ultrasound devices were available for 406 of 407 participants in the sonographer-acquired test set, and 112 of 114 participants in the novice-acquired test set. Sweeps collected with the low cost device were available for 104 of 407 participants in the sonographer-acquired test set, and 56 of 114 participants in the novice-acquired test set. Analyzable data from the low cost device became available later during the study, and this group of patients is representative of the full patient set. We randomly selected one study visit per patient for each analysis group to avoid combining correlated measurements from the same patient. For our fetal malpresentation model, the test set included 613 patients from the sonographer-acquired and novice-acquired datasets, resulting in 65 instances of non-cephalic presentation (10.6%). For each patient, the last study visit of the third trimester was included. Of note, there are more patients in the malpresentation model test set since the ground truth is not dependent on a prior visit. The disposition of study participants are summarized in STARD diagrams (Extended Data Figure 1) and Extended Data Table 1.

**Mobile-device-optimized AI gestational age and fetal malpresentation estimation**

We calculated the mean difference in absolute error between the GA model estimate and estimated gestational age as determined by standard fetal biometry measurements using imaging from traditional ultrasound devices operated by sonographers.[26] The reference ground truth GA was established as described above (Figure 1A). When conducting pairwise statistical comparisons between blind sweep and standard fetal biometry absolute errors, we established an a priori criterion for non-inferiority which was confirmed if the blind sweep mean absolute error (MAE) was less than 1.0 day greater than the standard fetal biometry's MAE. Statistical estimates and comparisons were computed after randomly selecting one study visit per patient for each analysis group, to avoid combining correlated measurements from the same patient.

We conducted a supplemental analysis of GA model prediction error with mixed effects regression on all test data, combining sonographer-acquired and novice-acquired test sets. Fixed effect terms accounted for the ground truth GA, the type of ultrasound machine used (standard vs. low cost), and the training level of the ultrasound operator (sonographer vs. novice). All patient studies were included in the analysis, and random effects terms accounted for intra-patient and intra-study effects.

GA analysis results are summarized in Table 1. The MAE for the GA model estimate with blind sweeps collected by sonographers using standard ultrasound devices was significantly lower than the MAE for the standard fetal biometry estimates (mean difference -1.4 ± 4.5 days, 95% CI -1.8, -0.9 days). There was a trend towards increasing error for bind sweep and standard fetal biometry procedures with gestational week (Figure 2, top left).

The accuracy of the fetal malpresentation model for predicting non-cephalic fetal presentation from third trimester blind sweeps was assessed using a reference standard determined by sonographers equipped with traditional ultrasound imagery (described above). We selected the latest study visit in the third trimester for each patient. Data from sweeps performed by the sonographers and novices were analyzed separately. We evaluated the fetal malpresentation model's area under the receiver operating curve (AUC-ROC) on the test set in addition to non-cephalic sensitivity and specificity.

The fetal malpresentation model attained an AUC-ROC of 0.977 (95% CI 0.949, 1.00), sensitivity of 0.938 (95% CI 0.848, 0.983), and specificity of 0.973 (95% CI 0.955, 0.985) (Table 2 and Figure 3).

**Generalization of GA and malpresentation estimation to novices**
Our models were trained on up to 15 blind sweeps per study performed by sonographers. No novice-acquired blind sweeps were used to train our models. We assessed GA model generalization to blind sweeps performed by novice operators that performed 6 sweeps. We compared the MAE between novice-performed blind sweep AI estimates and the standard fetal biometry. For the malpresentation model, we reported the AUC-ROC for blind sweeps performed by novices, along with the sensitivity and specificity at the same operating point used for evaluating blind sweeps performed by sonographers.

In this novice-acquired dataset, the difference in MAE between blind sweep AI estimates and the standard fetal biometry was -0.6 days (95% CI -1.7, 0.5), indicating that sweeps performed by novice operators provide a non-inferior GA estimate compared to the standard fetal biometry. Table 1 provides novice blind sweep performance analyzed by ultrasound device type. The mixed effects regression error analysis did not indicate a significant association between GA error magnitude and the type of operator conducting the blind sweep (P = .119).

Fetal malpresentation using novice-acquired blind sweeps was compared to the sonographer's determination on 189 participants (21 malpresentations), and AUC-ROC was 0.992 (95% CI 0.983, 1.0). On the preselected operating point, sensitivity was 1.0 (95% CI 0.839, 1.0) and specificity was 0.952 (95% CI 0.908, 0.979).

**Performance of low cost ultrasound device in GA and fetal malpresentation estimation**
GA model estimation using blind sweeps acquired with the low cost ultrasound device were compared against the clinical standard on the combined novice-acquired and sonographer-acquired test sets. We used the same a priori criterion for non-inferiority as described above, 1.0 day. For the malpresentation model, we reported AUC-ROC by type of ultrasound device along with sensitivity and specificity at the same operating point discussed above.

GA model estimation using blind sweeps acquired with the low cost ultrasound device were compared against the standard fetal biometry estimates on the combined novice-acquired and sonographer-acquired test sets. The blind sweep AI system had MAE of 3.98 ± 3.54 days versus 4.17 ± 3.74 days for standard fetal biometry (mean difference -0.21 ± 4.21, 95% CI -0.87, 0.44) which meets the criterion for non-inferiority.

Paired GA estimates for blind sweeps acquired with both a standard ultrasound device and the low cost device were available for some study participants in the combined test set (N=155 participants). The MAE difference between blind sweeps performed with the low cost and standard devices was 0.45 days (95% CI, 0.0, 0.9). The mixed effects regression showed that use of the low-cost device was associated with increased error magnitude (P = 0.001), although the estimated effect was only 0.67 days.

Fetal malpresentation estimation using blind sweeps acquired with the low cost ultrasound device were compared against the sonographer's determination on the combined novice-acquired and sonographer-acquired test sets (213 participants, 29 malpresentations). The blind sweep AI system had AUC-ROC of 0.97 (95% CI 0.944, 0.997). At the preselected operating point, sensitivity was 0.931 (95% CI 0.772, 0.992) and specificity was 0.94 (95% CI 0.896, 0.970).

**Simplified sweep evaluation**
Protocols consisting of fewer sweeps than the standard 6 sweeps (Figure 1B) may simplify clinical deployment. We selected M and R sweep types as the best performing set of two sweeps on the tuning set and evaluated this reduced protocol on the test sets.

On test set sweeps performed by sonographers, the reduced protocol of just the M and R sweep types (Figure 1B) was sufficient for maintaining non-inferiority of the blind sweep protocol relative to the standard fetal biometry estimates (MAE difference 95% CI: [-1.5, -0.69] days). The reduced protocol was sufficient for maintaining non-inferiority of blind sweeps relative to standard fetal biometry on test set examinations performed by novices (MAE difference 95% CI: [-1.19, 0.67] days). On average, the reduced protocol can be completed in 20.1 seconds, as extrapolated from videos collected from novices (see Extended Data Table 2). MAE across subgroups using the reduced protocol are provided in Table 1 (last row).

**Feedback Score Evaluation**
Our GA model provided a feedback score to evaluate the suitability of a video sequence for GA estimation. The GA model computed the feedback score for 24 frame video sequences (about one second in length) and therefore provided a semi-continuous feedback signal across the duration of a typical ten second long blind sweep. The feedback score took the form of an inverse variance estimate and can be used to weight and aggregate GA predictions across blind sweep video sequences during a study visit. All GA results were computed using this inverse variance weighting aggregation method. More details are provided in the Methods section.

As expected, video sequences with high feedback score had low MAE when compared against ground truth GA, and low feedback score video sequences had high MAE compared against ground truth GA. Figure 2 (bottom right) indicates the calibration of the feedback score on the held out test datasets. Extended Data Figure 2C shows example blind sweep video frames from high and low feedback score video sequences. The feedback score qualitatively aligns with the degree to which the fetus is visible in the video clip, with the high feedback score left and center-left examples showing fetal abdomen and head (respectively). In contrast, the fetus is not visible in the low feedback score examples (center-right and right).

**Runtime evaluation on mobile phones**
Our blind sweep AI models were designed to enable near real time inference on modern mobile phones or standard computers, to enable elimination of waiting time during the clinical examination procedure. We measured both GA and fetal malpresentation model run time performance using ten second long blind sweep videos, which was chosen to match the average length of novice blind sweeps (Extended Data Table 2). Videos were streamed to an Android test application running on Google Pixel 3 and 4, Samsung Galaxy S10, and Xiaomi Mi 9 phones (examples of Android phones that can be purchased refurbished for less than $250 USD). Videos were processed by the GA and fetal malpresentation models simultaneously, with both models executed in the same TensorFlow Lite runtime environment. All necessary image preprocessing operations were also included in the benchmark.

Our results indicated that combined diagnoses for both models are available between 0.2 and 1.0 seconds on average after the completion of a blind sweep on devices with a graphics processing unit (GPU), and between 1.5 and 2.5 seconds on average after completion on devices with neural network acceleration libraries for standard CPU processors. See Table 2 for complete benchmark results.

**Discussion**

In this study, we demonstrated how our end-to-end blind sweep mobile-device-optimized AI models can assist novices in LMICs in acquiring blind sweep ultrasounds to estimate two important obstetric measurements: GA and fetal malpresentation. While there have been multiple GA models proposed in the past, ours is the first to describe an end-to-end system focusing on use in LMIC settings. Three prior studies have described using deep learning on single video frames to either directly estimate GA or estimate head circumference that is then used in fetal biometry formulas.[21–24] One prior study describes using the FAMLI dataset to estimate GA through deep learning of blind sweep ultrasound videos.[24] Our model performs as well or better than previously described models. Our GA model estimation was non-inferior to standard fetal biometry estimates and our fetal malpresentation model had high sensitivity and specificity. Both models also had similar performance across sonographer and novice acquired ultrasounds.

We found that later in pregnancy there was less deterioration of GA model estimation accuracy compared to the clinical standard fetal biometry. Our models utilize the entire ultrasound video as opposed to only accounting for isolated biometric measurements (e.g. head circumference, femur length..) per the standard fetal biometry. This holistic approach may account for the increased accuracy later in the pregnancy, when the correlation between GA and physical size of the fetus is less pronounced. This may be especially helpful in providing more accurate estimated GA in LMIC settings where access to ultrasound in early pregnancy is rare.[27]

While pairwise comparisons between traditional devices and low cost devices suggest that traditional devices may perform slightly better, GA model estimation from low cost devices was non-inferior to standard fetal biometry estimation. This suggests that variation in device performance does not result in clinically significant differences in GA model performance. For our fetal malpresentation model, performance was similar between low cost and traditional devices.

We focused on improving user experience and simplifying ultrasound acquisition since some of the most vulnerable patients are in geographically remote areas without Internet access. While we initially evaluated on blind sweeps consisting of 6 sweeps, we found that our GA model performed similarly using only 2 sweeps. The compatibility of the GA model with this simplified 2-sweep protocol suggests that we may be able to simplify acquisition complexity for novices. Our GA model generates a real-time feedback score that provides information on ultrasound video quality and reliability for use in our AI models. In a clinical setting, this feedback score can potentially notify the ultrasound operator to redo a poorly performed sweep. Both the GA and malpresentation models along with the video quality feedback score have been optimized to run on affordable mobile devices and do not require Internet access.

One limitation of this study is the small sample size, which makes it difficult to evaluate each subgroup individually: novices, sonographers, and ultrasound device type. Our dataset included very few videos for GA less than 12 weeks and greater than 37 weeks so we cannot ensure the AI models generalize for these groups. In addition, we only had a limited number of non-cephalic presentations resulting in wide confidence intervals. We plan to validate our findings on a larger cohort to address these limitations. These future studies will also include predictions for other critical maternal fetal diagnostics and pregnancy risk stratification.

While designing the AI models, we addressed obstacles that may be encountered in low-resourced settings where remote care is often delivered through novices with limited training. Overall, tools such as the ones assessed in this study can potentially assist in upleveling the capabilities of both facility and novices in providing more advanced antenatal care. Additionally, the underlying techniques and technology could be applied and studied in other ultrasound-based clinical workflows. Prospective clinical evaluation and on other datasets will be important to evaluate real-world effectiveness and adaptations may be needed to integrate tools such as this into real-world workflows.[28]

**Tables and figures**

**Figure 1: Development of an AI system to acquire and interpret blind sweep ultrasound for antenatal diagnostics. A:** Datasets were curated from sites in Zambia and the USA and include ultrasound acquired by sonographers and midwives. Ground truth for gestational age was derived from the initial exam as part of clinical practice. An AI system was trained to identify gestational age and fetal malpresentation, and was evaluated by comparing the accuracy of AI predictions with the accuracy of clinical standard procedures. The AI system was developed using only sonographer blind sweep data and its generalization to novice users was tested on midwife data. Design of the AI system considered suitability for LMIC deployments in three ways: first, the system interpreted ultrasound from low-cost portable ultrasound devices; second, near real time interpretation is available offline on mobile phone devices; and finally, the AI system produces feedback scores that can be used to provide feedback to users. **B**: Blind sweep ultrasound acquisition procedure. The procedure can be performed by novices with a few hours of ultrasound training. While the complete protocol involves six sweeps, a set of two sweeps (M and R) were found to be sufficient for maintaining accuracy of gestational age estimation.

**A**

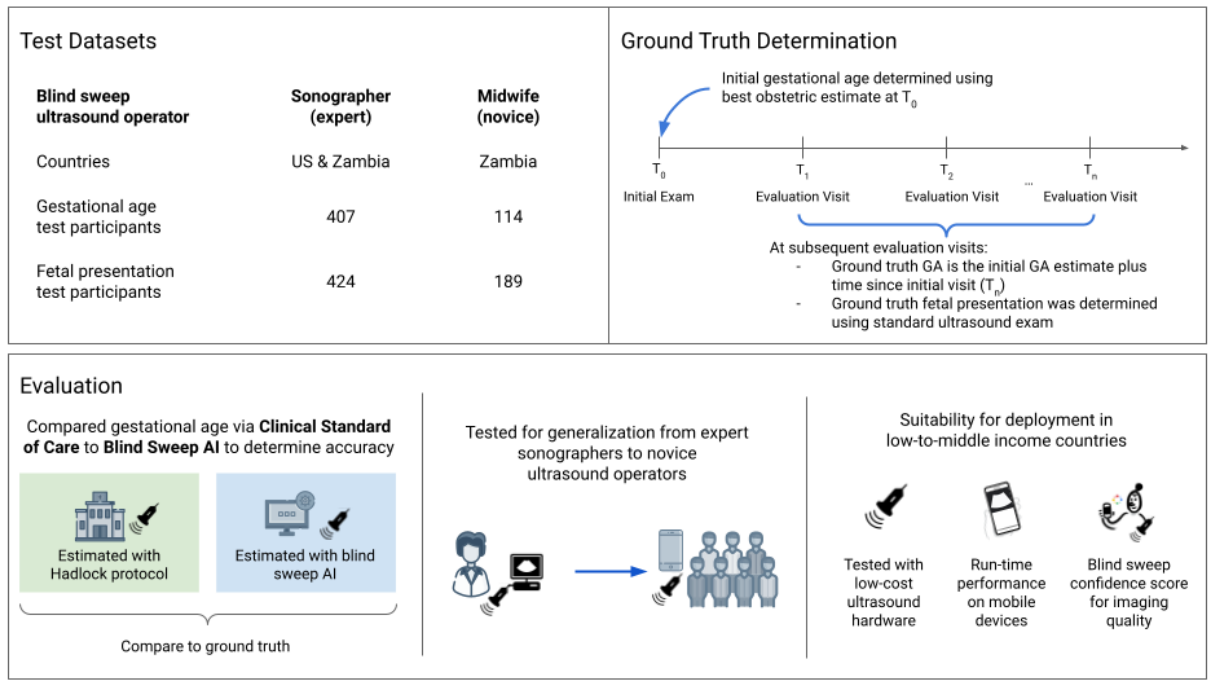

**B**

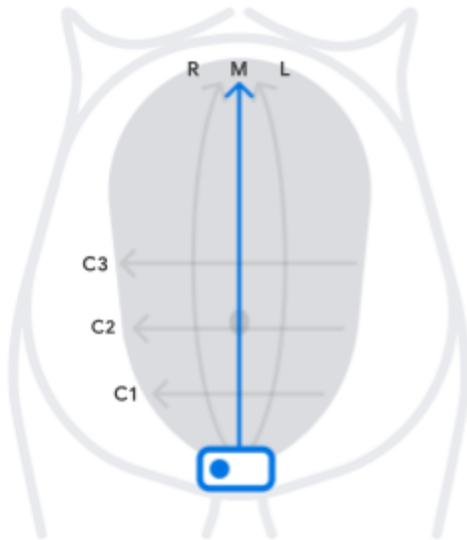

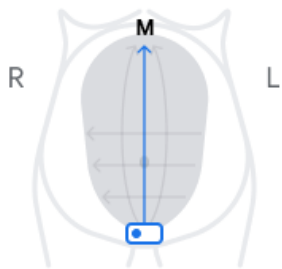
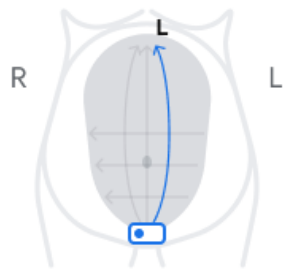
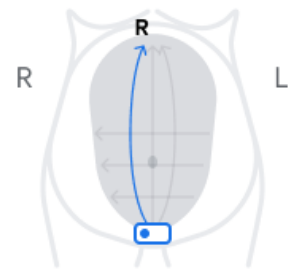

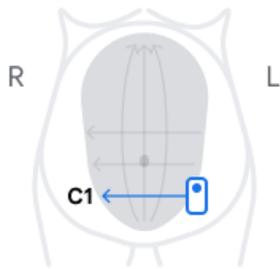
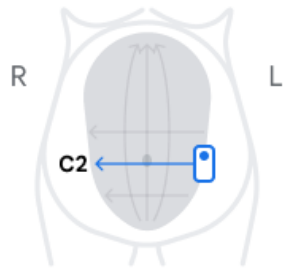
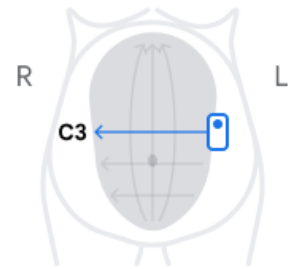

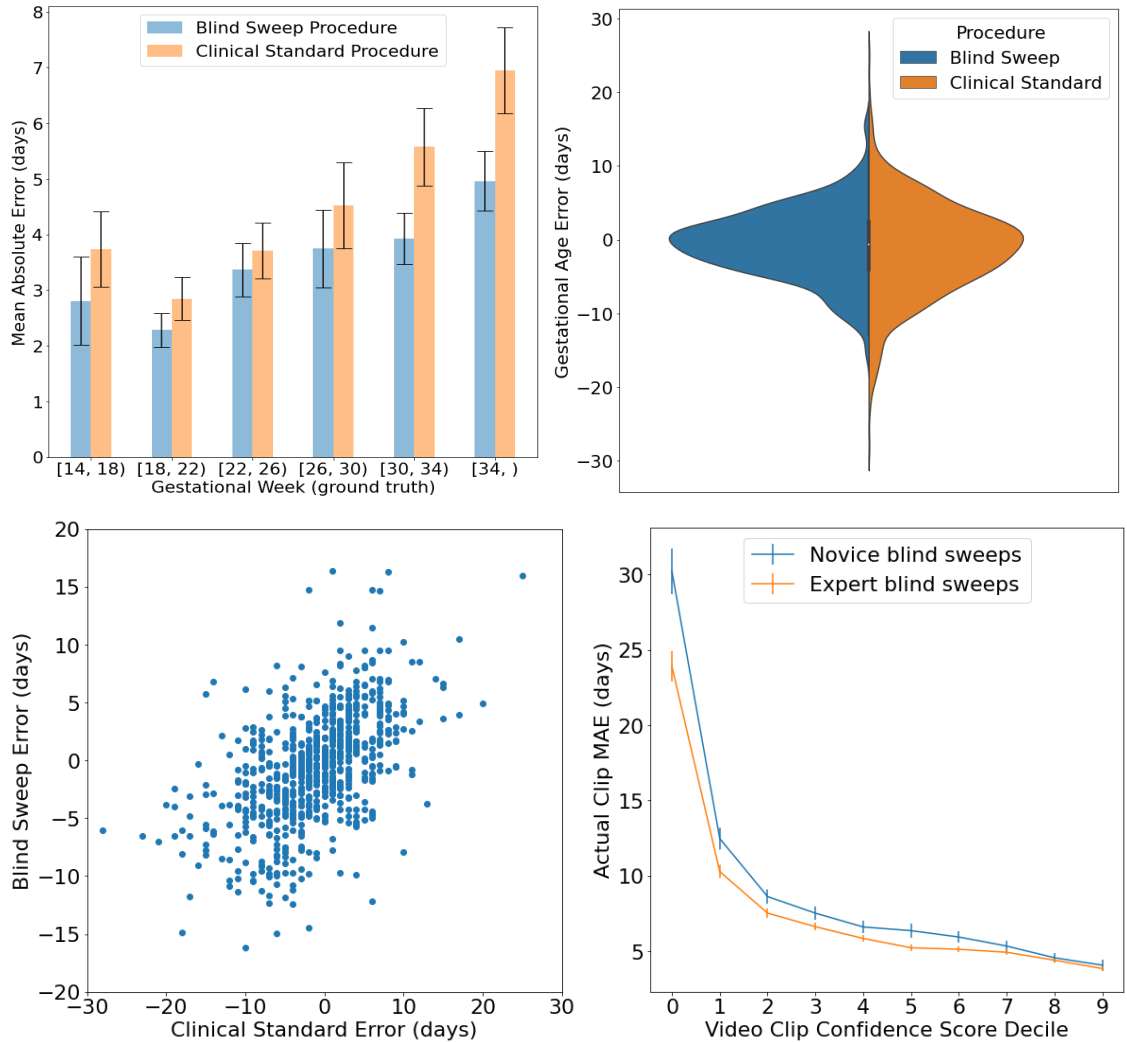

**Figure 2: Gestational Age Estimation. Top Right:** Error distributions for blind sweep procedure and standard fetal biometry procedure. **Bottom Left:** Paired errors for blind sweep and standard fetal biometry estimates in the same study visit. The errors of the two methods exhibit correlation, but the worst case errors for the blind sweep procedure have lower magnitude than the standard fetal biometry method. **Bottom Right:** Video sequence feedback score calibration on the test sets. The realized model estimation error on held out video sequences decreases as the model's feedback score increases. A thresholded feedback score may be used as a user feedback signal to redo low quality blind sweeps.

**Figure 3: Fetal Malpresentation Estimation:** Receiver operating characteristic (ROC) curves for fetal malpresentation estimation. Crosses indicate the predefined operating point selected from the tuning data set.

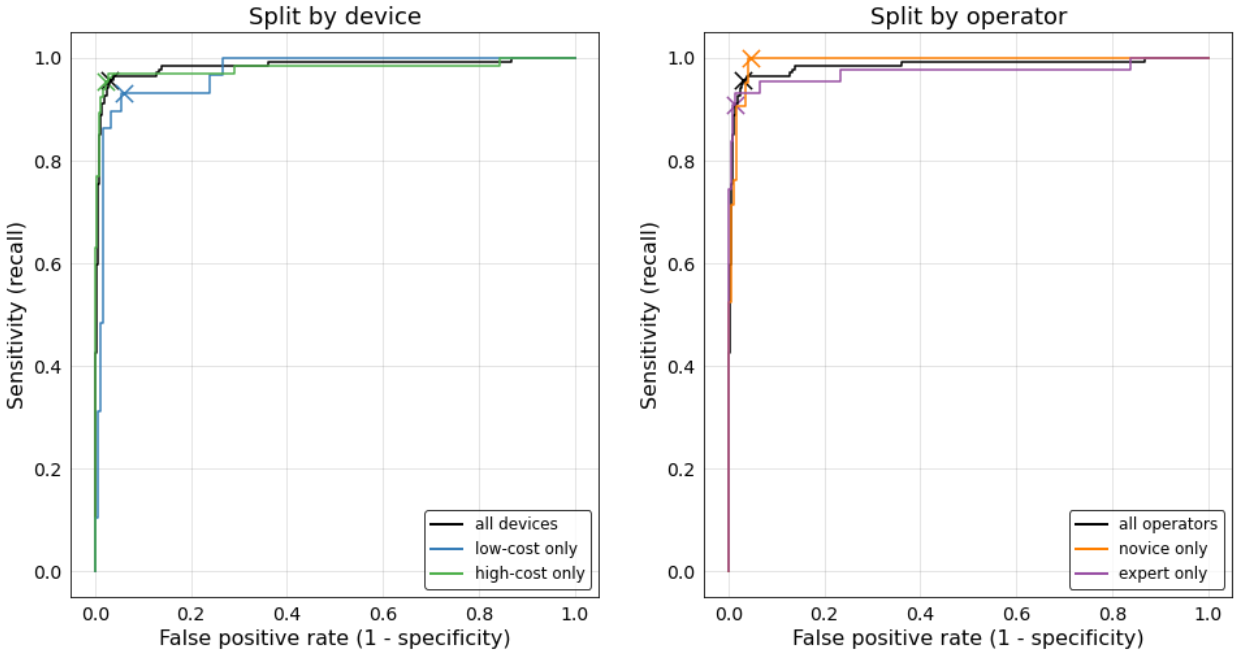

**Table 1: Gestational Age Estimation.** Mean absolute error (MAE) and mean error (ME) between GA estimated using the blind sweep procedure and ground truth, and the MAE and ME between the GA estimated using the standard fetal biometry ultrasound procedure and ground truth. One visit by each participant eligible for each subgroup was selected at random. The reduced blind sweep protocol (last row) included only two blind sweeps. All other blind sweep results used a set of 6 blind sweeps per patient visit.

|  | Data collected by sonographers | | Data collected by novices | |
| --- | --- | --- | --- | --- |
|  | Standard ultrasound device | Low cost handheld device | Standard ultrasound device | Low cost handheld device |
| Number | 406 | 104 | 112 | 56 |
| Blind sweep MAE ± sd (days) | 3.8 ± 3.6 | 3.3 ± 2.8 | 4.4 ± 3.5 | 5.0 ± 4.0 |
| Standard fetal biometry estimates MAE ± sd (days) | 5.2 ± 4.6 | 3.8 ± 3.6 | 4.8 ± 3.7 | 4.7 ± 4.0 |
| Blind sweep - Standard fetal biometry mean difference ± sd (days) | -1.4 ± 4.5 | -0.6 ± 3.8 | -0.4 ± 4.8 | 0.4 ± 5.1 |
| MAE difference 95% CI (days) | -1.8, -0.9 | -1.3, 0.1 | -1.3, 0.5 | -1.0, 1.7 |
| Blind sweep ME ± sd (days) | -0.9 ± 5.3 | 0.4 ± 4.4 | -1.5 ± 5.5 | -3.8 ± 5.4 |
| Standard fetal biometry estimates ME ± sd (days) | -1.4 ± 7.0 | -0.25 ± 5.4 | -2.6 ± 5.3 | -3.4 ± 5.2 |
| Reduced blind sweep protocol MAE ± sd (days) | 4.0 ± 3.7 | 3.5 ± 3.0 | 4.5 ± 3.5 | 5.1 ± 4.2 |

**Table 2: Fetal Malpresentation Estimation.** The fetal malpresentation model was assessed by comparing predictions to the determination of a sonographer. In each subset of the data, we selected only the latest eligible visit from each patient. For sensitivity and specificity computations, model predictions were binarized according to a predefined threshold. Confidence intervals on the area under the receiver operating characteristic (AUC-ROC) were computed using the DeLong method. Confidence intervals on sensitivity and specificity were computed with the Clopper-Pearson method.

| Subset | Number of Participants | Number of Malpresentations | AUC-ROC (95% CI) | Sensitivity (95% CI) | Specificity (95% CI) |
|---|---|---|---|---|---|
| **All** | 613 | 65 | 0.977 (0.949, 1.0) | 0.938 (0.848, 0.983) | 0.973 (0.955, 0.985) |
| **Low-cost only** | 213 | 29 | 0.970 (0.944, 0.997) | 0.931 (0.772, 0.992) | 0.940 (0.896, 0.970) |
| **High-cost only** | 598 | 65 | 0.980 (0.953, 1.000) | 0.954 (0.871, 0.990) | 0.977 (0.961, 0.988) |
| **Novice only** | 189 | 21 | 0.992 (0.983, 1.000) | 1.000 (0.839, 1.000) | 0.952 (0.908, 0.979) |
| **Sonographer only** | 424 | 43 | 0.972 (0.933, 989) | 0.907 (0.779, 0.974) | 0.987 (0.970, 0.996) |

**Table 3: Mobile Device Model Runtime Benchmarks.** Time to model inference results (mean and standard deviation in seconds) measured from the end of a ten second long blind sweep video. Both gestational age and fetal malpresentation models run simultaneously on the same video sequence and image preprocessing operations are included. Near real time inference is achievable on smart phones with graphics processor units or compute libraries optimized for neural network operations. This enables a simple and fast examination procedure in clinical environments.

| Mobile phone | Processor Type | | |
| --- | --- | --- | --- |
| | GPU mean ± standard deviation | CPU w/ XNNPACK library (4 threads) | CPU (4 threads) |
| Pixel 3 | 0.9 ± 0.1 s | 2.1 ± 1.0 s | 13.2 ± 2.9 s |
| Pixel 4 | 0.2 ± 0.1 s | 1.5 ± 0.8 s | 9.8 ± 2.5 s |
| Samsung Galaxy S10 | 0.5 ± 0.1 s | 1.7 ± 1.1 s | 10.3 ± 2.3 s |
| Xiaomi Mi 9 | 1.0 ± 0.2 s | 1.8 ± 1.3 s | 13.7 ± 3.4 s |

**Competing Interests and Funding**
This study was partially funded by Google Inc. Above authors with Google, Inc. affiliation are employees and own stock as part of the standard employee compensation package. This study was partially funded by the Bill and Melinda Gates Foundation (OPP1191684, INV003266). The conclusions and opinions expressed in this article are those of the authors and do not necessarily reflect those of the Bill and Melinda Gates Foundation.

**Acknowledgments**
We would like to thank Yun Liu and Cameron Chen for helpful feedback on the manuscript, and Tiya Tiyasirichokchai for help with figure illustrations.

## Methods

**Algorithm Development**

We developed two deep learning neural network models to predict GA and fetal malpresentation. Our models generated diagnostic predictions directly from ultrasound video: sequences of image pixel values were the input and an estimate of the clinical quantity of interest was the output. The GA model produced an estimate of age, measured in days, for each blind sweep video sequence. The GA model additionally provided an estimate of its confidence in the estimate for a given video sequence, see Model Architecture section below for details. No intermediate fetal biometric measurements were required during training or generated during inference. The fetal malpresentation model predicted a probability score between 0.0 and 1.0 for whether the fetus is in non-cephalic presentation.

In the USA, the ground truth GA was determined for each participant based on the "best obstetric estimate," either part of routine clinical care, using procedures recommended by the American College of Obstetricians and Gynecologists (ACOG) [29]. The best obstetric estimate combines information from the last menstrual period (LMP), GA derived from assisted reproductive technology (if applicable), and fetal ultrasound anatomic measurements. In Zambia, only the first fetal ultrasound was used to determine the ground truth GA as the LMP in this setting was considered less reliable as patients often presented for care later in pregnancy.

The GA model was trained on sonographer-acquired blind sweeps (up to 15 sweeps per patient) as well as sonographer-acquired "fly-to" videos that capture five to ten seconds before the sonographer has acquired standard fetal biometry images. The fetal malpresentation model was only trained on blind sweeps. For each training set case, fetal malpresentation was specified as one of four possible values by a sonographer (cephalic, breech, transverse, oblique), and dichotomized to "cephalic" vs "non-cephalic".

Our analysis cohort included all pregnant women in the FAMLI and Novice User Study datasets who had the necessary ground truth information for gestational age and fetal presentation from September 2018 to January 2021. Study participants were assigned at random to one of 3 dataset splits: train, tune, or test. We used the following proportions: 60% train / 20% tune / 20% test for study participants who did not receive novice sweeps, and 10% tune / 90% test for participants who received novice sweeps. The tuning set was used for optimizing machine learning training hyperparameters and selecting a classification threshold probability for the fetal malpresentation model. This threshold was chosen to yield equal non-cephalic specificity and sensitivity on the tuning set, blinded to the test sets. None of the blind sweep data collected by the novices were used for training.

Cases consisted of multiple blind sweep videos, and our models generated predictions independently for each video sequence within the case. For the GA model, each blind sweep was divided into multiple video sequences. For the fetal malpresentation model, video sequences corresponded to a single complete blind sweep. We then aggregated the predictions to generate a single case-level estimate for either GA or fetal malpresentation (described further in the Model Architecture section below).

**Model Architecture**
Our deep learning models consist of convolutional and recurrent components (Extended Data Figure 3). The convolutional component uses the MobileNetV2 network [30] as a feature extractor for each video frame. The final feature layer of MobileNetV2 generates a sequence of image embeddings which are then processed by a grouped convolutional LSTM cell[31]. Since the recurrent connections in the model only operate on condensed features extracted from individual frames by MobileNetV2 and don't require the original frames, the forward pass requires relatively little memory, helping to enable deployment on mobile devices. LSTM state and output spatial feature maps are spatially average-pooled and transformed via a fully connected layer before final sigmoid, linear, or softplus units (depending on the prediction task) to compute the model's predictions. The model's final prediction is the output of these units after processing the last frame in the video sequence.

We specialized this architecture for two distinct diagnostic models: gestational age prediction and fetal malpresentation screening. We defined gestational age prediction as a regression problem in which the model produces an estimate of gestational age, measured in days, for each video sequence. The model operates on log-transformed labels and uses linear output units. To make predictions, we exponentiate the raw model output to recover the original scale. The gestational age model also provides an estimate of expected variance for each video sequence, which is generated by an additional softplus model output. This expected variance is inverted to give the confidence feedback score discussed in the main text. We use the mean-variance regression loss function proposed in [32,33] to jointly train the model's log-transformed age and expected variance outputs. The combined loss function has the following form:

$$\sum_i \frac{(f(x_i) - y)^2}{g(x_i)} + \log g(x_i)$$

where $f(x_i)$ is the gestational "log-age" prediction for video clip i, y is the (log-transformed) age label, and $g(x_i)$ is the model's estimate of expected variance for the clip.

The fetal malpresentation model predicts a binary value indicating whether the fetus is in cephalic (head down) versus non-cephalic presentation. This output is produced by a sigmoid unit and the model is trained with a standard binary cross-entropy classification loss function.

**Data Preprocessing**
Ultrasound videos have variable physical scale-- the anatomical size associated with image pixels varies according to depth (frequency) settings of the ultrasound hardware device. The physical scale value, typically in units of centimeters per pixel, is calculated by the ultrasound device and provided as metadata along with recorded video. Hardware depth settings are fixed during blind sweeps and therefore the physical scale value is constant throughout the duration of each video clip, but may vary substantially across patient studies. We denote the video's physical scale as α$_i$ below.

We found that model performance improves dramatically in the tune set when video frames are rescaled to a constant scale value α across the training set. The rescaling procedure first uses bilinear interpolation to resize the number of pixels in each frame according to $\alpha_i/\alpha$, which normalizes the per-pixel physical scale. Then the resized image is cropped or padded to fixed height and width. The scale constant α, height, and width were chosen by visually inspecting training set images to ensure that relevant ultrasound image details were not excessively cropped, but otherwise these values were not extensively tuned. The same image rescaling procedure is performed as a pre-processing step during model inference. The gestational age model benefits from high resolution images, and we used 576 x 432 pixels with α=0.0333 centimeters/pixel. The fetal malpresentation model operates with lower resolution, and we used 320 x 240 pixels with α=0.06 centimeters/pixel.

Blind sweep video sequences are divided into multiple equal length clips that correspond to a fixed LSTM sequence length, and each clip serves as an independent training example. We also apply temporal subsampling in order to reduce computational load during inference and to ensure that clips capture enough context across the span of the sweep motion. For the gestational age model, we use ½ temporal subsampling and clips of 24 frames. We find that aggregating predictions from multiple short clips works well for gestational age estimation (see Mobile Inference section below for details on our aggregation method). The fetal malpresentation model benefits from longer frame sequences that capture the full duration of the blind sweep video. We use ⅓ temporal subsampling and 100 frame clips in order to provide the model with enough spatial context regarding the presentation of the fetus.

**Training**
The gestational age regression model uses the gestational age ground truth associated with the case as the training label for all video clips within the case. We use both blind sweeps and biometry fly-to videos from the FAMLI dataset during gestational age training. The fetal malpresentation model uses only second and third trimester cases, since fetal presentation is not well defined during early pregnancy. We restrict training to blind sweep videos only. For each training set case, fetal presentation is specified as one of four possible values by an expert sonographer (cephalic, breech, transverse, oblique.) We transform this to a binary training label by grouping breech, transverse, and oblique presentations into a single non-cephalic category.

The MobileNetV2 feature extractor's weights were pre-trained on ImageNet[34] data, and further refined along with the rest of the model weights during training on ultrasound data. Training was done using TensorFlow running on third generation tensor processing units with a 4x4 topology. During training we applied random data augmentation: horizontal and vertical flips as well as random image crops are uniformly applied across all frames within the training video clip. The LSTM state and output tensors have channel width equal to 512. We applied dropout with a keep probability of 0.863 for the gestational age model and 0.8 for the fetal malpresentation model. Model parameters were learned via AdamW optimization [35] with a batch size of 8.

Learning rates were decreased from an initial starting point according to a linear ramp based on the training step number. The gestational age model began with a learning rate of 4.58e-4 and ended with 4.58e-7 after 1 million training steps. The fetal presentation model began with learning rate 3.14e-5 and ended with 3.14e-7 after 300,000 training steps. Dropout probabilities and learning rate schedules were optimized based on tuning set performance.

**Mobile Device Inference**

Our trained models were converted to a suitable format for deployment on mobile devices using the TensorFlow Lite (TF-Lite) framework. After training in TensorFlow, we replaced the input layer in the model graph with a TF-Lite compatible input layer. The TF-Lite model performs inference on a single image at each time step, and accepts the current image and previous LSTM state from the previous time step as inputs.

During inference we used the same image re-scaling and temporal subsampling settings that are used in training, but we omitted random data augmentation. All reported model evaluation results were produced by running inference with the converted model in a TF-Lite runtime, which provides identical results independent of the underlying hardware used for evaluation.

Each case contains multiple blind sweep videos of varying length. We divided them into equal length video clips that match the LSTM sequence length used during training. We used the model's prediction on the final frame of the sequence as the single prediction for the clip. Predictions for all video clips in each case were aggregated together to generate a single prediction for the case. The gestational age model uses an inverse variance weighting procedure [36] to combine the clip-level predictions $x_c$, into a case-level mean gestational age $\bar{x}$ using estimated variance $\sigma_c$ of each clip

$$\bar{x} = \frac{\sum_c x_c/\sigma_c^2}{\sum_c 1/\sigma_c^2}$$

Since the model fits log-transformed gestational age $f = log(GA)$ and the variances $g$ are fitted to a normal distribution of $f$, the true variance of gestational age $\sigma_c$ needs correction using the log-normal distribution variance formula

$$\sigma_c = [\exp(g^2) - 1] \exp(2\mu + g^2)$$

where the mean μ = $f_c$.

The fetal presentation model generates one prediction per blind sweep video, and we average the model's sigmoid probability outputs to generate a case-level probability that the fetus is in non-cephalic presentation.

We selected our model architecture to optimize for runtime performance on standard mobile devices, such that the model can perform real-time inference on a live ultrasound video stream. We explored model weight quantization and inference delegate configurations, but settled on non-quantized floating point model weights since performance was sufficient when using modern mobile phones with graphics processing units or neural network acceleration libraries.

**Extended Data Figure 1: STARD Diagrams. A:** Blind sweeps performed by trained sonographers (FAMLI study). **B:** Blind sweeps performed by midwives (novice ultrasound operators.)

**A**

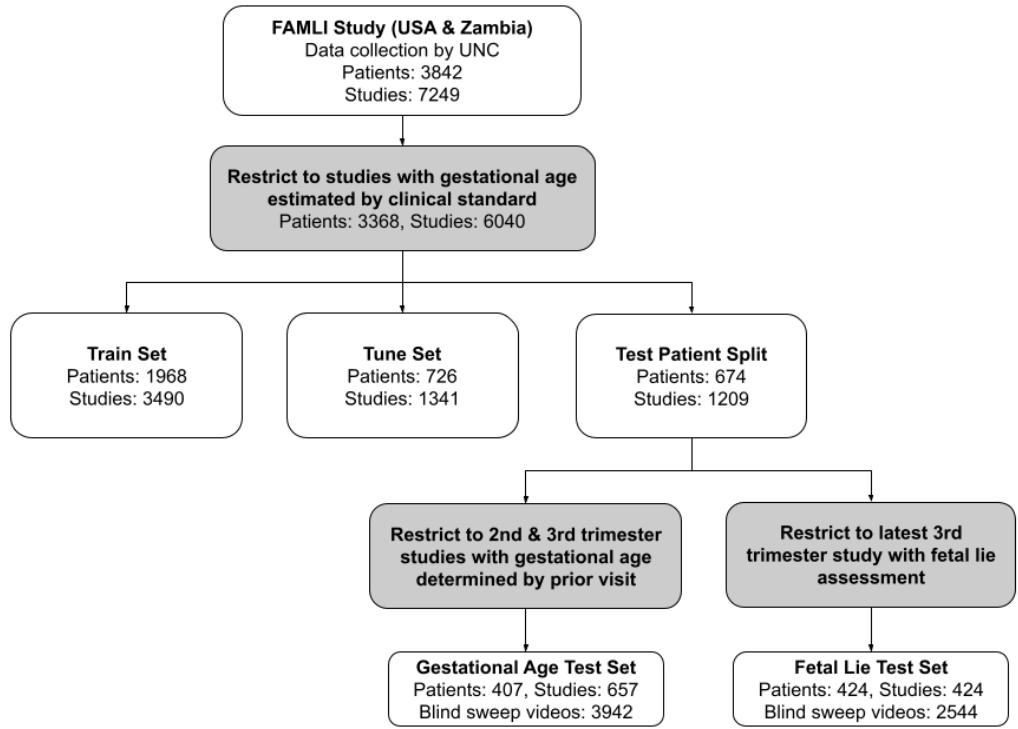

**B**

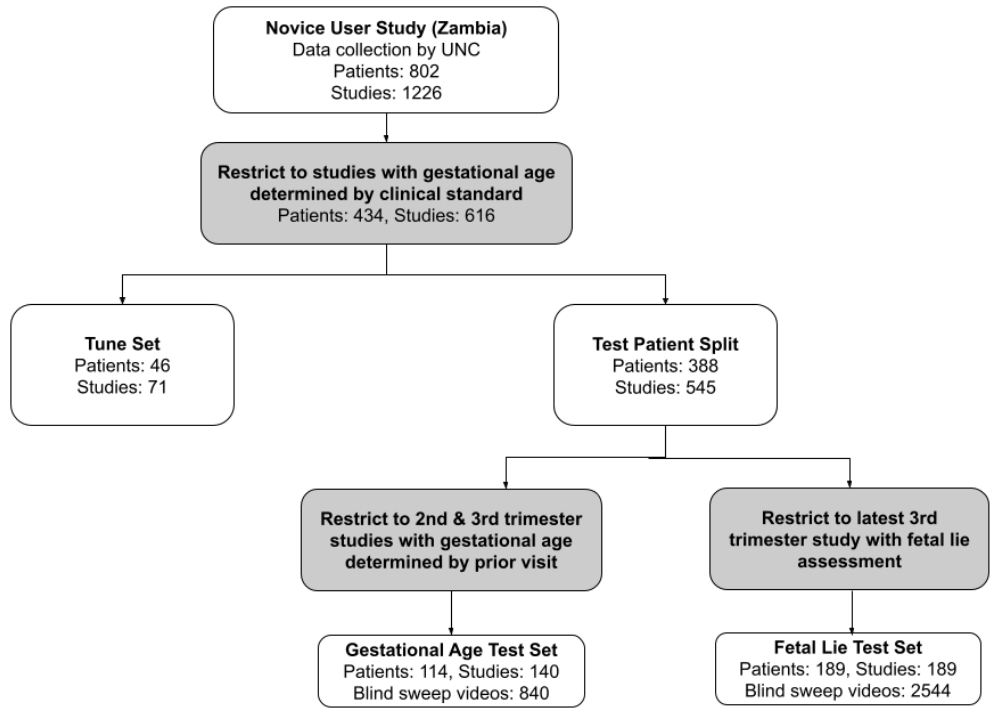

**Extended Data Figure 2: Blind sweep and Biometry Images. A, left:** Abdominal circumference measured by a sonographer. The clinical standard for estimating gestational age requires measurements of fetal anatomy captured from precise standardized viewpoints. **A, center & right:** Blind sweep example video frames. Blind sweep AI models do not require standardized anatomical views and videos may be acquired by ultrasound operators with little training. **B, left:** Example video frame acquired by a high end ultrasound device (GE Voluson family). See also row A for examples from SonoSite M-Turbo. **B, center, right:** Example video frames acquired by a low cost portable ultrasound device (Butterfly IQ). **C, left & center left:** Examples of blind sweep video frames with high feedback score for gestational age model. The fetal abdomen (left) and head (center left) are visible. **C, center right & right:** Examples of blind sweep video frames with low feedback score for gestational age model. The fetus is not visible in either example.

**A**

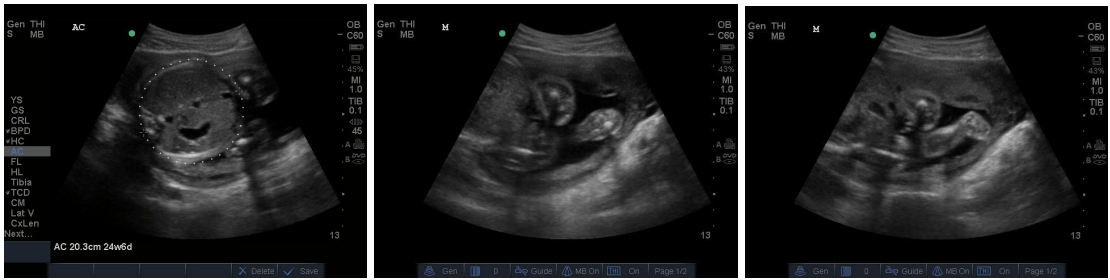

**B**

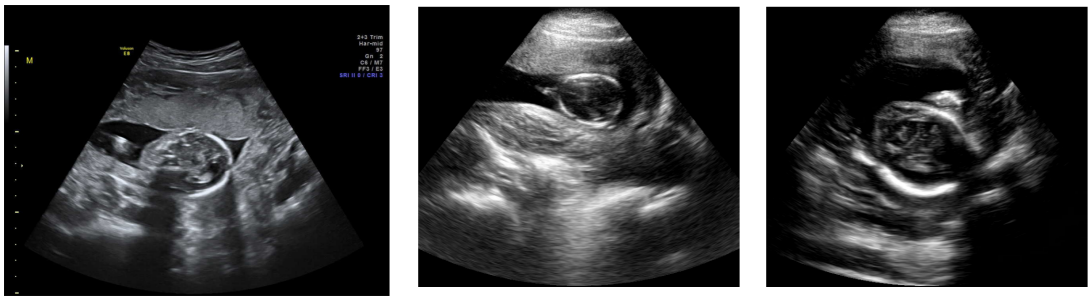

**C**

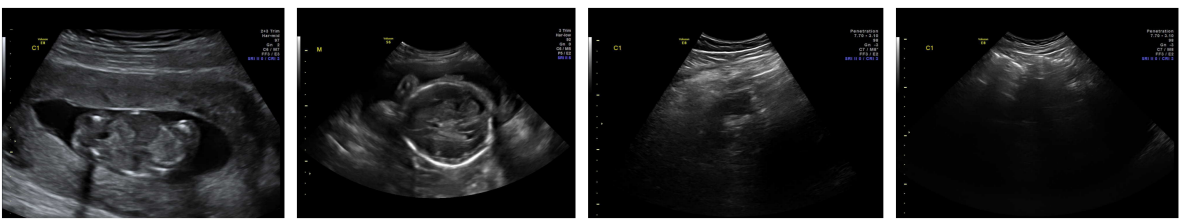

**Extended Data Figure 3: Artificial Intelligence Model Architecture.** The MobileNetV2 network [30] is used as a feature extractor applied to each video frame. The final feature layer of MobileNetV2 generates a sequence of image embeddings which is then processed by the grouped convolutional LSTM cell[31]. This architecture was selected for its suitability for deployment on mobile phones.

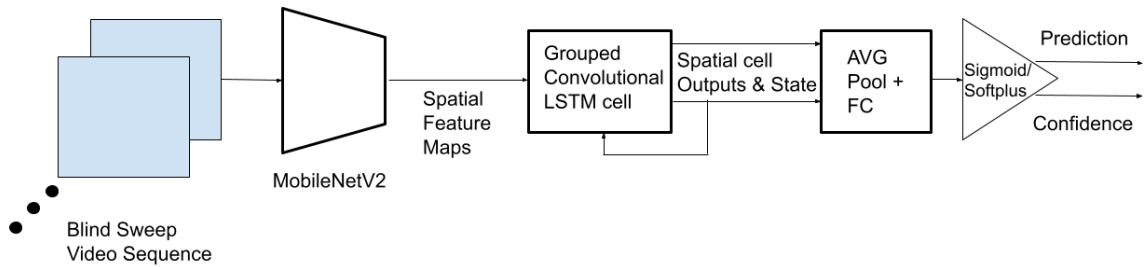

**Extended Data Table 1: Characteristics of study participants.**

| Variable | Sonographer group (n=407) | Novice group (n=114) |
|---|---|---|
| Participant age at enrollment (± standard deviation) | 28.8 ± 5.6 years | 28.0 ± 5.8 years |
| Gestational age at first study visit (± standard deviation) | 159.4 ± 49.7 days | 167.8 ± 47.4 days |
| Total number of study visits during second trimester (%) | 273 (41.6%) | 40 (28.8%) |
| Total number of study visits during third trimester | 384 (58.5%) | 99 (71.2%) |
| Fetal malpresentation (%) | 10.3% | 11.1% |

**Extended Data Table 2: Average Length of Blind Sweeps Performed by Novices.** See Figure 1B for a depiction of the blind sweep types.

| Blind sweep type | Mean Length ± standard deviation |
| --- | --- |
| M | 10.0 ± 4.2 seconds |
| R | 10.1 ± 4.1 seconds |
| L | 10.2 ± 4.2 seconds |
| C1 | 9.6 ± 3.8 seconds |
| C2 | 9.7 ± 4.6 seconds |
| C3 | 9.3 ± 3.5 seconds |